\documentclass[10pt,journal,compsoc]{IEEEtran}

\usepackage{color}

\usepackage[english]{babel}

\usepackage{graphicx}
\usepackage{caption}
\usepackage{subcaption}
\usepackage{wrapfig}
\usepackage{hyperref}

\usepackage{booktabs}
\usepackage{makecell}

\usepackage{algorithm}
\usepackage[noend]{algpseudocode}
\usepackage{varwidth}

\usepackage{etex}

\long\def\ignorethis#1{}

\def\equationautorefname~#1\null{Equation~(#1)\null}

\newcommand{\pixel}{d} %

\usepackage{xcolor}  %

\ifCLASSOPTIONcompsoc
  \usepackage[nocompress]{cite}
\else
  \usepackage{cite}
\fi
\ifCLASSINFOpdf
\else
\fi

\usepackage{amsmath}
\usepackage{amsfonts}
\usepackage{mathtools}

\usepackage{makecell}

\usepackage{babel}
\usepackage{cuted}

\begin{document}
\title{Visual Foresight: Model-Based Deep Reinforcement Learning for Vision-Based Robotic Control}

\author{Frederik Ebert*, Chelsea Finn*, Sudeep Dasari, Annie Xie, Alex Lee, Sergey Levine

\IEEEcompsocitemizethanks{
\IEEEcompsocthanksitem  The first two authors contributed equally.}%
\thanks{Manuscript received 11/22/2018}}

\markboth{Journal of \LaTeX\ Class Files,~Vol.~14, No.~8, August~2015}%
{Shell \MakeLowercase{\textit{et al.}}: Bare Demo of IEEEtran.cls for Computer Society Journals}
\IEEEtitleabstractindextext{%
\begin{abstract}
Deep reinforcement learning (RL) algorithms can learn complex robotic skills from raw sensory inputs, but have yet to achieve the kind of broad generalization and applicability demonstrated by deep learning methods in supervised domains. We present a deep RL method that is practical for real-world robotics tasks, such as robotic manipulation, and generalizes effectively to never-before-seen tasks and objects. In these settings, ground truth reward signals are typically unavailable, and we therefore propose a self-supervised model-based approach, where a predictive model learns to directly predict the future from raw sensory readings, such as camera images. At test time, we explore three distinct goal specification methods: designated pixels, where a user specifies desired object manipulation tasks by selecting particular pixels in an image and corresponding goal positions, goal images, where the desired goal state is specified with an image, and image classifiers, which define spaces of goal states. Our deep predictive models are trained using data collected autonomously and continuously by a robot interacting with hundreds of objects, without human supervision. We demonstrate that visual MPC can generalize to never-before-seen objects---both rigid and deformable---and solve a range of user-defined object manipulation tasks using the same model.
\end{abstract}

}

\maketitle
\begin{strip}
     \centering\noindent
     \includegraphics[width=\textwidth]{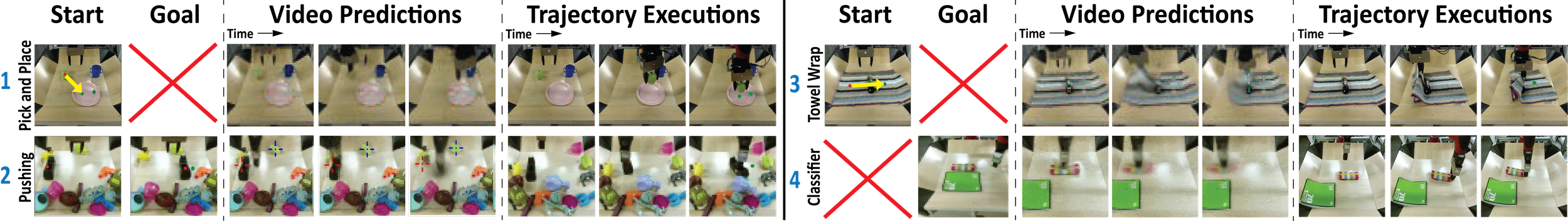}
	\captionof{figure}{Our approach trains a single model from unsupervised interaction that generalizes to a wide range of tasks and objects, while allowing flexibility in goal specification and both rigid and deformable objects not seen during training. Each row shows an example trajectory. From left to right, we show the task definition, the video predictions for the planned actions, and the actual executions.
	Tasks can be defined as (top) moving pixels corresponding to objects, (bottom left) providing a goal image, or (bottom right) providing a few example goals.
	Best viewed in PDF. \\ \\}
	\label{fig:example_traj}
\end{strip}
\vspace{10mm}

\IEEEdisplaynontitleabstractindextext

\IEEEpeerreviewmaketitle

\IEEEraisesectionheading{\section{Introduction}\label{sec:introduction}}

Humans are faced with a stream of high-dimensional sensory inputs and minimal external supervision, and yet, are able to learn a range of complex, generalizable skills and behaviors.
While there has been significant progress in developing deep reinforcement learning algorithms that learn complex skills and scale to high-dimensional observation spaces, such as pixels~\cite{tdgammon,atari,e2e,alphago}, learning behaviors that \emph{generalize} to new tasks and objects remains an open problem.
The key to generalization is diversity. When deployed in a narrow, closed-world environment, a reinforcement learning algorithm will recover skills that are successful only in a narrow range of settings. 
Learning skills in diverse  environments, such as the real world, presents a number of significant challenges: external reward feedback is extremely sparse or non-existent, and the agent has only indirect access to the state of the world through its senses, which, in the case of a robot, might correspond to cameras and joint encoders.

We approach the problem of learning generalizable behavior in the real world from the standpoint of sensory prediction. Prediction is often considered a fundamental component of intelligence \cite{predictionandbrain}. Through prediction, it is possible to learn useful concepts about the world even from a raw stream of sensory observations, such as images from a camera. If we predict raw sensory observations directly, we do not need to assume availability of low-dimensional state information or an extrinsic reward signal.  Image observations are both information-rich and high-dimensional, presenting both an opportunity and a challenge. Future observations provide a substantial amount of supervisory information for a machine learning algorithm. However, the predictive model must have the capacity to predict these high-dimensional observations, and the control algorithm must be able to use such a model to effectively select actions to accomplish human-specified goals.  Examples of such goals are shown in figure \ref{fig:example_traj}.

We study control via prediction in the context of robotic manipulation, formulating a model-based reinforcement learning approach centered around prediction of raw sensory observations. One of the biggest challenges in learning-based robotic manipulation is generalization: how can we learn models that are useful not just for a narrow range of tasks seen during training, but that can be used to perform new tasks with new objects that were not seen previously?
Collecting a training dataset that is sufficiently rich and diverse is often challenging in highly-structured robotics experiments, which depend on human intervention for reward signals, resets, and safety constraints. We instead set up a minimally structured robotic control domain, where data is collected by the robot via unsupervised interaction with a wide range of objects, making it practical to collect large amounts of interaction data. The robot collects a stream of raw sensory observations (image pixels), without any reward signal at training time, and without the ability to reset the environment between episodes. This setting is both realistic and necessary for studying RL in diverse real-world environments, as it enables automated and unattended collection of diverse interaction experience. Since the training setting affords no readily accessible reward signal, learning by prediction presents an appealing option: the supervision signal for prediction is always available even in the stream of unsupervised experience.
We therefore propose to learn action-conditioned predictive models directly on raw pixel observations, and show that they can be used to accomplish a range of pixel-based manipulation tasks on a real robot in the physical world at test-time. 

\begin{figure}[t]
\centering
\includegraphics[width=\columnwidth,trim={0mm 0 0 0},clip]{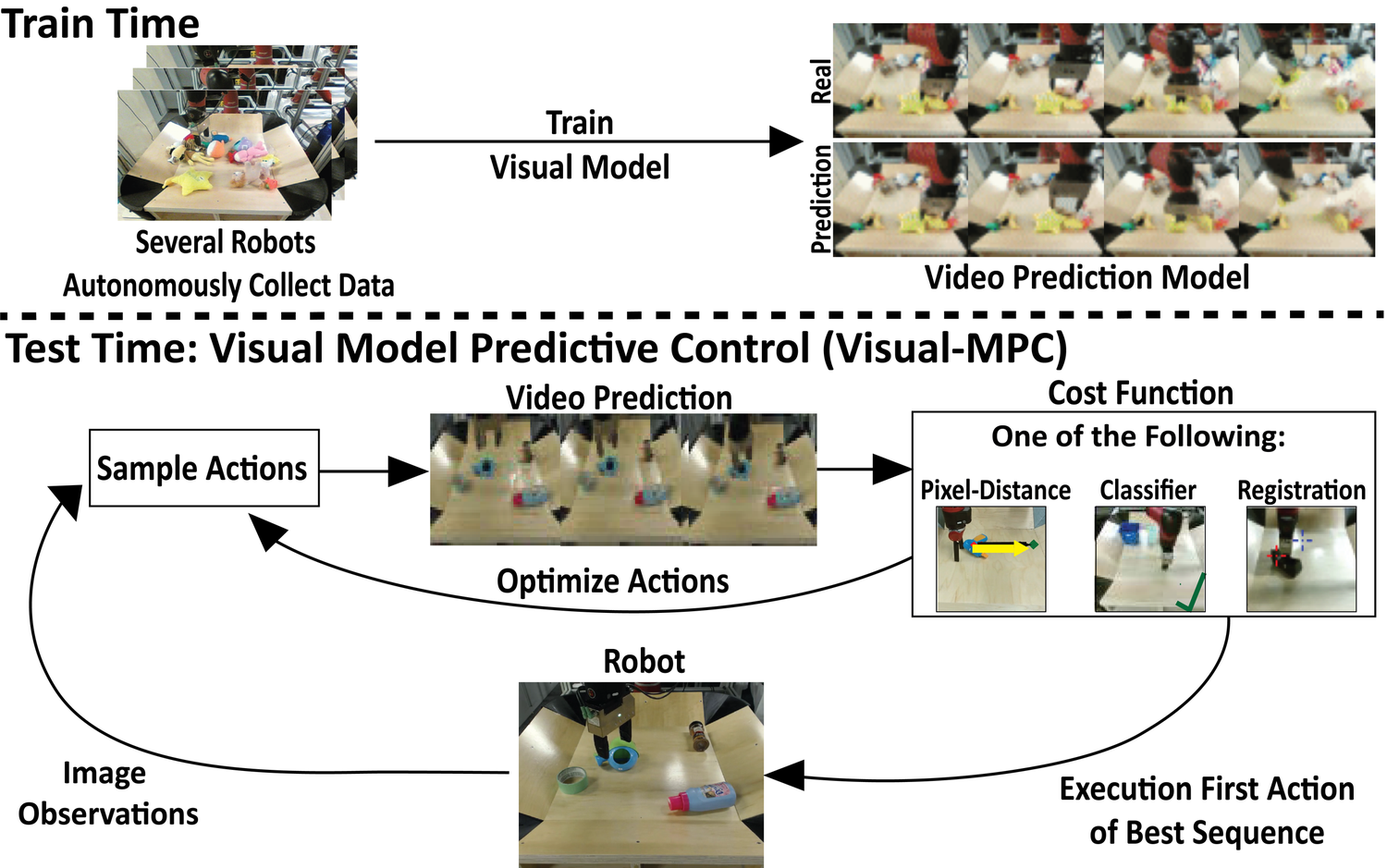}
\caption{\small{Overview of visual MPC. (top) At training time, interaction data is collected autonomously and used to train a video-prediction model. (bottom) At test time, this model is used for sampling-based planning. In this work we discuss three different choices for the planning objective.}}
\label{fig:overview}
\end{figure}

The main contributions of this work are as follows. We present \emph{visual MPC}, a general framework for deep reinforcement learning with sensory prediction models that is suitable for learning behaviors in diverse, open-world environments (see figure~\ref{fig:overview}).
We describe deep neural network architectures that are effective for predicting pixel-level observations amid occlusions and with novel objects. Unlike low-dimensional representations of state, specifying and evaluating the reward from pixel predictions at test-time is nontrivial: we present several practical methods for specifying and evaluating progress towards the goal---including distances to goal pixel positions, registration to goal images, and success classifiers---and compare their effectiveness and use-cases.
Finally, our evaluation shows how these components can be combined to enable a real robot to perform a range of object manipulation tasks from raw pixel observations. Our experiments include manipulation of previously unseen objects, handling multiple objects, pushing objects around obstructions, handling clutter, manipulating deformable objects such as cloth, recovering from large perturbations, and grasping and maneuvering objects to user-specified locations in 3D-space. Our results represent a significant advance in the \emph{generality} of skills that can be acquired by a real robot operating on raw pixel values using a single model.

This article combines and extends material from several prior conference papers~\cite{foresight,sna,ebert2018robustness,flo}, presenting them in the context of a unified system. We include additional experiments, including cloth manipulation and placing tasks, a quantitative \emph{multi-task} experiment assessing the performance of our method on a wide range of distinct tasks with a single model, as well as a comprehensive, open-sourced simulation environment to facilitate future research and better reproducibility. The code and videos can be found on the project webpage\footnote{For videos \& code: \url{https://sites.google.com/view/visualforesight}}.

\section{Related Work}\label{sec:rel_work}

\noindent \textbf{Model-based reinforcement learning.} Learning a model to predict the future, and then using this model to act, falls under the general umbrella of model-based reinforcement learning. Model-based RL algorithms are generally known to be more efficient than model-free methods~\cite{deisenroth2013survey}, and have been used with both low-dimensional~\cite{deisenroth2011pilco} and high-dimensional~\cite{deep_mpc} model classes. However, model-based RL methods that directly operate on raw image frames have not been studied as extensively. Several algorithms have been proposed for simple, synthetic images~\cite{watter2015embed} and video game environments~\cite{alexey, ha2018world, atarioh}, but have not been evaluated on generalization or in the real world, while other work has also studied model-based RL for individual robotic skills~\cite{dsae, zhang2018solar, se3}. In contrast to these works, we place special emphasis on \emph{generalization}, studying how predictive models can enable a real robot to manipulate previously unseen objects and solve new tasks.
Several prior works have also sought to learn inverse models that map from pairs of observations to actions, which can then be used greedily to carry out short-horizon tasks~\cite{agrawal2016learning,nair2017combining}. However, such methods do not directly construct longer-term plans, relying instead on greedy execution. In contrast, our method learns a forward model, which can be used to plan out a sequence of actions to achieve a user-specified goal.

\noindent \textbf{Self-supervised robotic learning.}
A number of recent works have studied self-supervised robotic learning, where large-scale unattended data collection is used to learn individual skills such as grasping~\cite{lerrel,google_handeye, calandra2017feeling,pinto2016curious}, push-grasp synergies~\cite{zeng2018learning}, or obstacle avoidance~\cite{greg_kahn_uncertainty,crashing}. 
In contrast to these methods, our approach learns predictive models that can be used to perform a variety of manipulation skills, and does not require a success measure, event indicator, or reward function during data collection.

\noindent \textbf{Sensory prediction models.}
We propose to leverage sensory prediction models, such as video-prediction models, to enable large-scale self-supervised learning of robotic skills. Prior work on action-conditioned video prediction has studied predicting synthetic video game images~\cite{atarioh,recurrentsimulators}, 3D point clouds~\cite{se3}, and real-world images~\cite{bootsetal,finn_nips,video_pixel_networks}, using both direct autoregressive frame prediction~\cite{beyond_mse,finn_nips,video_pixel_networks} and latent variable models~\cite{babaeizadeh2017stochastic,kurutach2018learning}. Several works have sought to use more complex distributions for future images, for example by using pixel autoregressive models~\cite{video_pixel_networks,scott_reed}. 
While this often produces sharp predictions, the resulting models are extremely demanding computationally.
Video prediction without actions has been studied for unstructured videos~\cite{beyond_mse,convlstm,vondrick} and driving~\cite{prednet,dynamic_filter_networks}.  In this work, we extend video prediction methods that are based on predicting a transformation from the previous image~\cite{finn_nips,dynamic_filter_networks}. 
\section{Overview}\label{sec:prelim}
\label{sec:vmpc}
In this section, we summarize our visual model-predictive control (MPC) method, which is a model-based reinforcement learning approach to end-to-end learning of robotic manipulation skills. Our method, outlined in Figure \ref{fig:overview}, consists of three phases: unsupervised data collection, predictive model training, and planning-based control via the model at test-time.

\noindent \textbf{Unsupervised data collection}: At \emph{training time}, data is collected autonomously by applying random actions sampled from a pre-specified distribution. It is important that this distribution allows the robot to visit parts of the state space that are relevant for solving the intended tasks. For some tasks, uniform random actions are sufficient, while for others, the design of the exploration strategy takes additional care, as detailed in Sections \ref{sec:system} and \ref{subsec:cloth_folding_data}.

\noindent \textbf{Model training}: Also during \emph{training time}, we train a video prediction model on the collected data. The model takes as input an image of the current timestep and a sequence of actions, and generates the corresponding sequence of future frames. This model is described in Section \ref{sec:model}.

\noindent \textbf{Test time control}: At \emph{test time}, we use a sampling-based, gradient free optimization procedure, similar to a shooting method~\cite{betts1998survey},
to find the sequence of actions that minimizes a cost function. Further details, including the motivation for this type of optimizer, can be found in Section \ref{sec:optimizer}. 

Depending on how the goal is specified, we use one of the following three cost functions. When the goal is provided by clicking on an object and a desired goal-position, a \emph{pixel-distance cost-function}, detailed in Section \ref{subsec:pixel_dist_cost}, evaluates how far the designated pixel is from the goal pixels. We can specify the goal more precisely by providing a goal image in addition to the pixel positions and make use of \emph{image-to-image registration} to compute a cost function, as discussed in Section \ref{subsec:reg_cost}. Finally, we show that we can specify more conceptual tasks
by providing one or several examples of success and employing a \emph{classifier-based} cost function as detailed in Section \ref{subsec:class_cost}. The strengths and weaknesses of different costs functions and trade-offs between them are discussed in Section \ref{subsec:cost_discuission}. 

The model is used to plan $T$ steps into the future, and the first action of the action sequence that attained lowest cost, is executed. In order to correct for mistakes made by the model, the actions are iteratively replanned at each real-world time step\footnote{With real-world step we mean timestep of the real-world as opposed to predicted timesteps.} $\tau \in \{0,...,\tau_{max}\}$ following the framework of model-predictive control (MPC). 
In the following sections, we explain the video-prediction model, the planning cost function, and the trajectory optimizer.

\section{Video Prediction for Control}
\label{sec:model}

\begin{figure}[t]
	\centering
	\includegraphics[width=0.9\columnwidth]{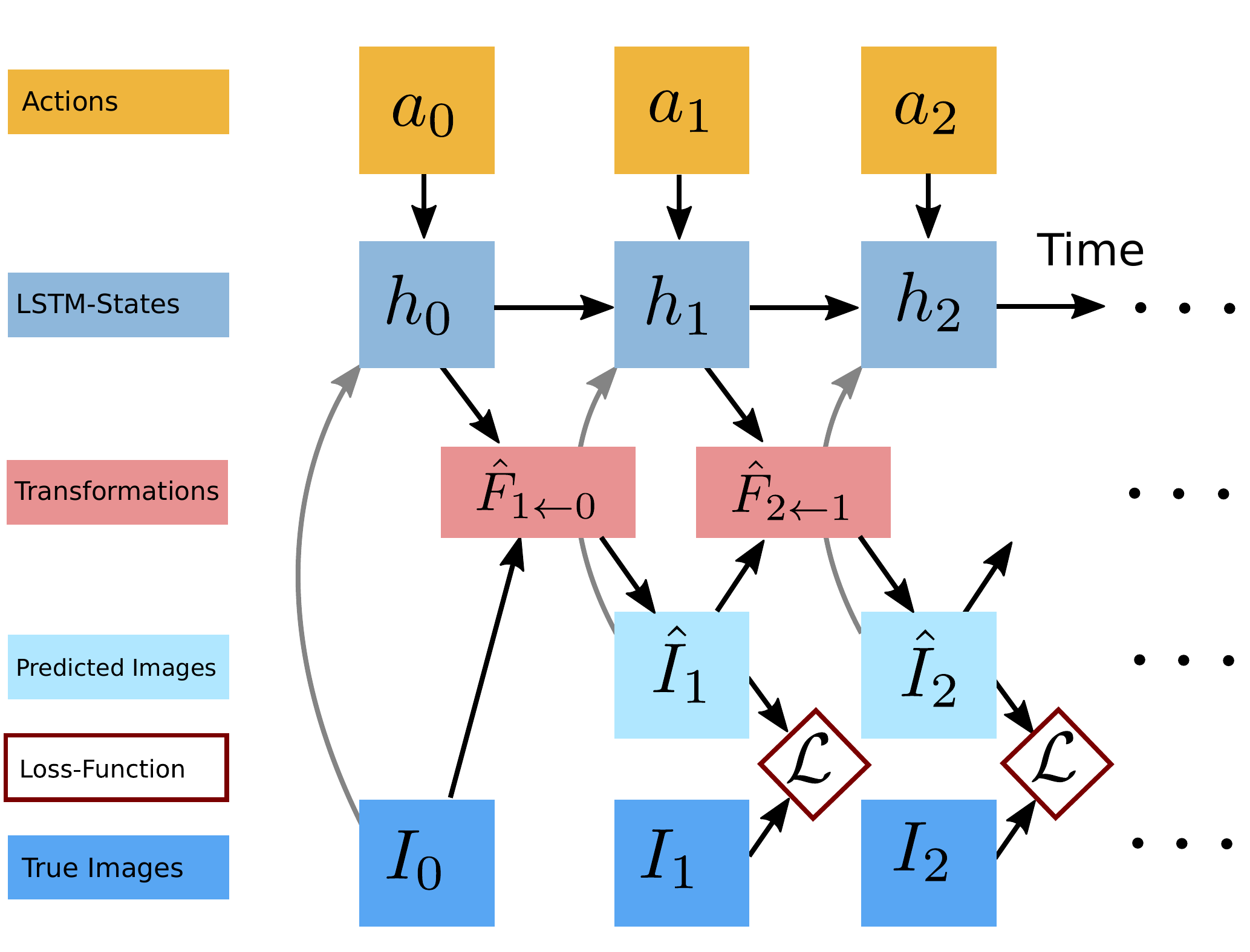}
	\caption{\small{Computation graph of the video-prediction model. Time goes from left to right, $a_t$ are the actions, $h_t$ are the hidden states in the recurrent neural network, $\hat{F}_{t+1 \leftarrow t}$ is a 2D-warping field, $I_t$ are real images, and $\hat{I}_t$ are predicted images, $\mathcal{L}$ is a pairwise training-loss.}}   
	\label{fig:prediction_model}
\end{figure}

In visual MPC, we use a transformation-based video prediction architecture, first proposed by Finn et al. \cite{finn_nips}. The advantage of using transformation-based models over a model that directly generates pixels is two-fold: (1) prediction is easier, since the appearance of objects and the background scene can be reused from previous frames and (2) the transformations can be leveraged to obtain predictions about where pixels will move, a property that is used in several of our planning cost function formulations. The model, which is implemented as a recurrent neural network (RNN) $g_{\theta}$ parameterized by $\theta$, has a hidden state $h_t$ and takes in a previous image and an action at each step of the rollout.  Future images $\hat{I}_{t+1}$ are generated by warping the previous generated image $\hat{I}_t$ or the previous true image $I_t$, when available, according to a 2-dimensional flow field $\hat{F}_{t+1 \leftarrow t}$. A simplified illustration of model's structure is given in figure \ref{fig:prediction_model}. It is also summarized in the following two equations:
\begin{align}
[h_{t+1}, \hat{F}_{t+1 \leftarrow t}] 	&= g_{\theta}(a_t, h_t, I_t) \\
\hat{I}_{t+1} 							&= \hat{F}_{t+1 \leftarrow t} \diamond  \hat{I}_t 
\label{simple_dna}
\end{align}
Here, the bilinear sampling operator $\diamond$ interpolates the pixel values bilinearly with respect to a location $(x,y)$ and its four neighbouring pixels in the image, similar to \cite{zhou2016view}. Note that, as shown in figure \ref{fig:prediction_model}, at the first time-step the real image is transformed, whereas at later timesteps previously generated images are transformed in order to generate multi-frame predictions. The model is trained with gradient descent on a $\ell_2$ image reconstruction loss, denoted by $\mathcal{L}$ in figure \ref{fig:prediction_model}.
\begin{figure}[t]
    \centering
    \includegraphics[width=\columnwidth]{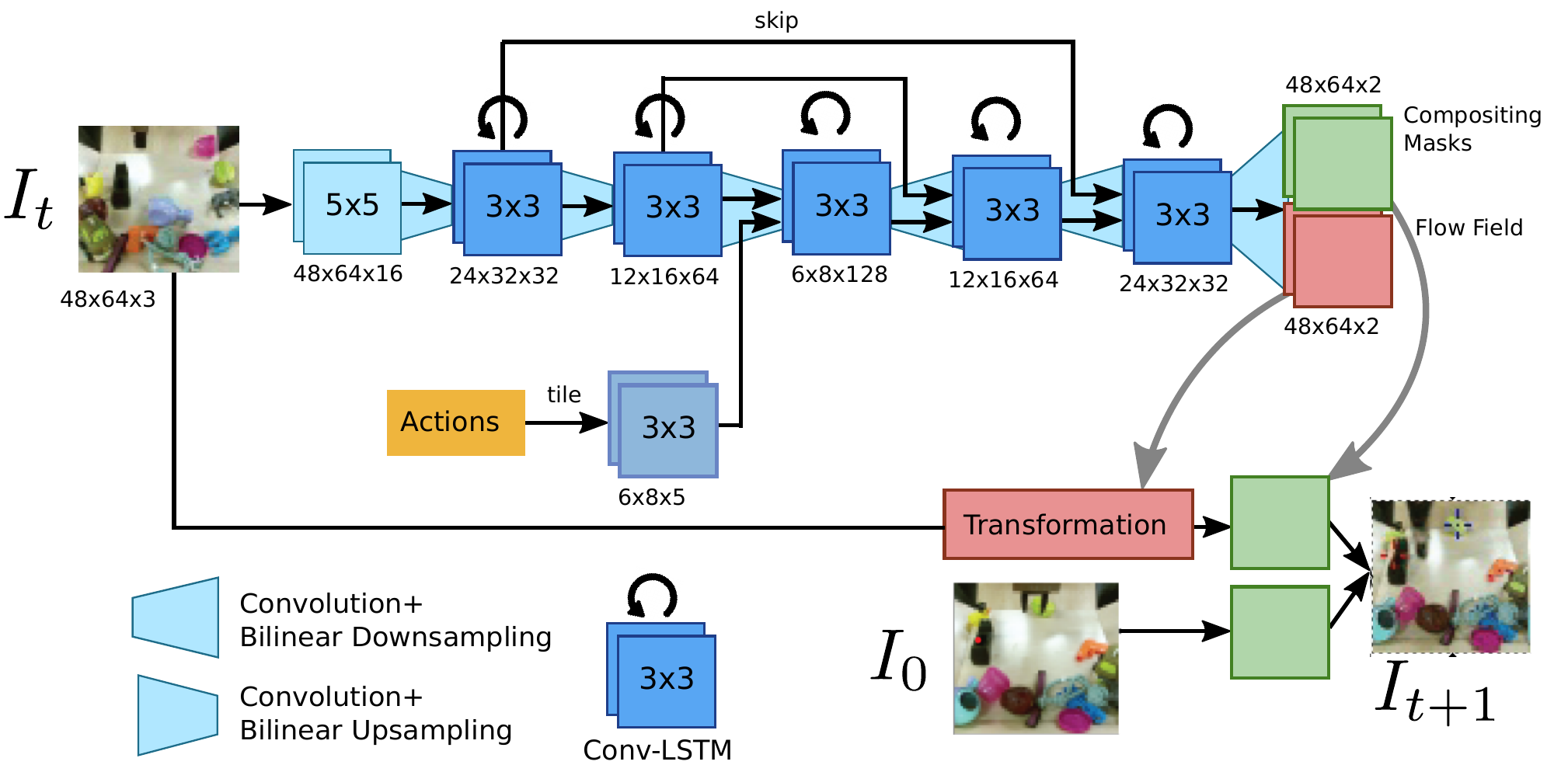}
    \caption{\small{Forward pass through the recurrent SNA model. The red arrow indicates where the image from the first time step $I_0$ is concatenated with the transformed images $\hat{F}_{t+1 \leftarrow t} \diamond  \hat{I}_t $ multiplying each channel with a separate mask to produce the predicted frame for step $t+1$.}}      \label{fig:occlusion_model}
\end{figure}
A forward pass of the RNN is illustrated in figure \ref{fig:occlusion_model}. We use a series of stacked convolutional LSTMs and standard convolutional layers interleaved with average-pooling and upsampling layers. The result of this computation is the 2 dimensional flow-field $\hat{F}_{t+1 \leftarrow t}$ which is used to transform a current image $I_t$ or $\hat{I}_t$. More details on the architecture are provided in Appendix \ref{sec:skipcon}.

\noindent \textbf{Predicting pixel motion.} 
When using visual MPC with a cost-function based on start and goal pixel positions, we require a model that can effectively predict the 2D motion of the user-selected start pixels $\pixel_0^{(1)}, \dots, \pixel_0^{(P)}$ up to $T$ steps into the future\footnote{Note that when using a classifier-based cost function, we do not require the model to output transformations.}.
More details about the cost functions are provided in section \ref{sec:cost}. Since the model we employ is transformation-based, this motion prediction capability emerges automatically, and therefore no external pixel motion supervision is required. To predict the future positions of the designated pixel $d$, the same transformations used to transform the images are applied to the distribution over designated pixel locations. The warping transformation $\hat{F}_{t+1 \leftarrow t}$ can be interpreted as a stochastic transition operator allowing us to make probabilistic predictions about future locations of individual pixels:
\begin{equation}
\hat{P}_{t+1} = \hat{F}_{t+1 \leftarrow t} \diamond  \hat{P}_t
\label{eqn:prob_forward}
\end{equation}
Here, $P_t$ is a distribution over image locations which has the same spatial dimension as the image. For simplicity in notation, we will use a single designated pixel moving forward, but using multiple is straightforward. At the first time step, the distribution $\hat{P}_0$ is defined as 1 at the position of the user-selected designated pixel and zero elsewhere. The distribution $\hat{P}_{t+1}$ is normalized at each prediction step.

Since this basic model, referred to as dynamic neural advection (DNA), predicts images only based on the previous image, it is unable to recover shapes (e.g., objects) after they have been occluded, for example by the robot arm. Hence, this model is only suitable for planning motions where the user-selected pixels are not occluded during the manipulation, limiting its use in cluttered environments or with multiple selected pixels. In the next section, we introduce an enhanced model, which lifts this limitation by employing temporal skip connections. 

\noindent \textbf{Skip connection neural advection model.}
To enable effective tracking of objects through occlusions, we can add temporal skip connections to the model: we now transform pixels not only from the previously generated image $\hat{I}_t$, but from all previous images $\hat{I}_1,...\hat{I}_{t}$, including the context image $I_0$, which is a real image. All these transformed images can be combined to a form the predicted image $\hat{I}_{t+1}$ by taking a weighted sum over all transformed images, where the weights are given by masks $\mathbf{M}_t$ with the same size as the image and a single channel:
\begin{equation}
\hat{I}_{t+1} =  \mathbf{M}_{0} (\hat{F}_{t+1 \leftarrow 0} \diamond I_t) +  \sum_{j=1}^{\tau} \mathbf{M}_{j} (\hat{F}_{t+1 \leftarrow j} \diamond  \hat{I}_j).
\end{equation}
We refer to this model as the \emph{skip connection neural advection model (SNA)}, since it handles occlusions by using temporal skip connections such that when a pixel is occluded, e.g., by the robot arm or by another object, it can still reappear later in the sequence.
Transforming from all previous images comes with increased computational cost, since the number of masks and transformations scales with the number of time-steps $\tau$. However, we found that in practice a greatly simplified version of this model, where transformations are applied only to the previous image and the \emph{first image} of the sequence $I_0$, works equally well. Moreover we found that transforming the first image of the sequence is not necessary, as the model uses these pixels primarily to generate the image background. Therefore, we can use the first image directly, without transformation. More details can be found in the appendix \ref{sec:skipcon} and \cite{sna}. %

\section{Planning Cost Functions}
\label{sec:cost}

In this section, we discuss how to specify and evaluate goals for planning. One na\"{i}ve approach is to use pixel-wise error, such as $\ell_2$ error, between a \emph{goal image} and the \emph{predicted image}. However there is a severe issue with this approach: large objects in the image, i.e. the arm and shadows, dominate such a cost; therefore a common failure mode occurs when the planner matches the arm position with its position in the goal image, disregarding smaller objects. This failure motivates our use of more sophisticated mechanisms for specifying goals, which we discuss next.

\subsection{Pixel Distance Cost}
\label{subsec:pixel_dist_cost}
A convenient way to define a robot task is by choosing one or more \emph{designated pixels} in the robot's camera view and choosing a destination where each pixel should be moved. For example, the user might select a pixel on an object and ask the robot to move it 10 cm to the left. This type of objective is general, in that it can define any object relocation task on the viewing plane. Further, success can be measured quantitatively, as detailed in section \ref{sec:experiments}.
Given a distribution over pixel positions $P_0$, our model predicts distributions over its positions $P_t$ at time $t \in \{ 1, \dots, T \}$. One way of defining the cost per time-step $c_t$ is by using the expected Euclidean distance to the goal point $d_g$, which is straight-forward to calculate from $P_t$ and $g$, as follows:
 \begin{align}
c = \sum_{t = 1, \dots, T} c_t =  \sum_{t = 1, \dots, T} \mathbb{E}_{\hat{d}_{t} \sim P_{t}} \left[\|\hat{d}_{t} - d_{g}\|_2\right] 
 \label{eq:cost}
 \end{align}
The per time-step costs $c_t$ are summed together giving the overall planing objective $c$. The expected distance to the goal provides a smooth planning objective and enables longer-horizon tasks, since this cost function encourages movement of the designated objects into the right direction for each step of the execution, regardless of whether the goal-position can be reached within $T$ time steps or not. This cost also makes use of the uncertainty estimates of the predictor when computing the expected distance to the goal. For multi-objective tasks with multiple designated pixels $d^{(i)}$ the costs are summed to together, and optionally weighted according to a scheme discussed in \autoref{subsec:reg_cost}.  

\subsection{Registration-Based Cost}
\label{subsec:reg_cost}
We now propose an improvement over using pixel distances. When using pixel distance cost functions, it is necessary to know the \emph{current} location of the object,  $\pixel_0^{(1)}, \dots, \pixel_0^{(P)}$ at each replanning step, so that the model can predict the positions of this pixel from the current step forward.
To update the belief of where the target object currently is, we propose to register the current image to the start and optionally also to a \emph{goal image}, where the designated pixels are marked by the user. Adding a goal image can make visual MPC more precise, since when the target object is close to the goal position, registration to the goal-image greatly improves the position estimate of the designated pixel. Crucially, the registration method we introduce is self-supervised, using the same exact data for training the video prediction model and for training the registration model. This allows both models to continuously improve as the robot collects more data.

\begin{figure}
	\centering
	\includegraphics[width=0.7\linewidth]{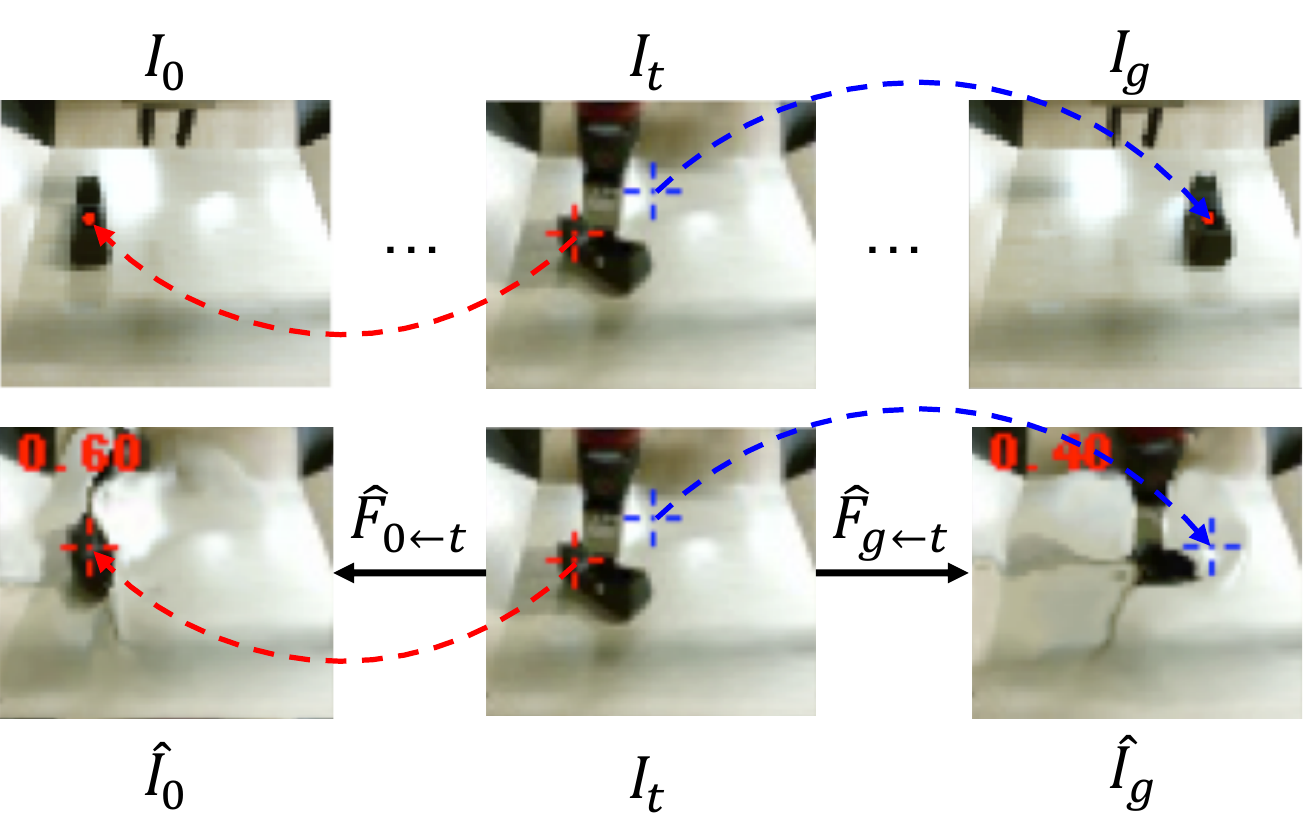}
	\vspace{-0.3cm}
	\caption{\small{Closed loop control is achieved by registering the current image $I_t$ globally to the first frame $I_0$ and the goal image $I_g$. In this example registration to $I_0$ succeeds while registration to $I_g$ fails since the object in $I_g$ is too far away.}
		\label{fig:reg_single}
	}
\end{figure}

\begin{figure}[t!]
	\centering
	\begin{subfigure}[b]{0.35\linewidth}
		\centering
		\includegraphics[width=\linewidth]{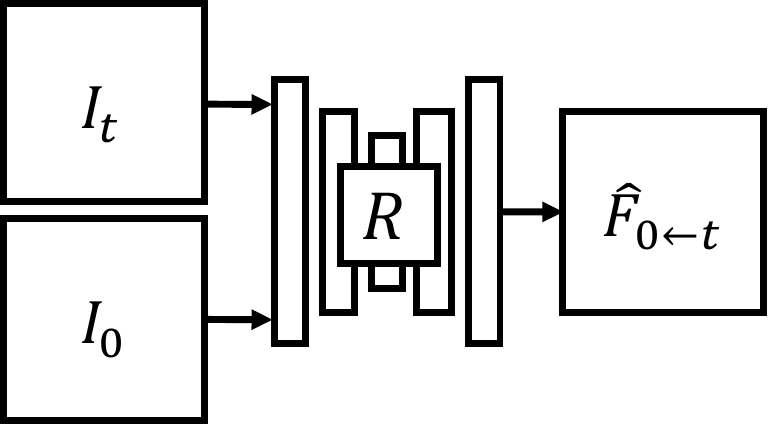}\vspace{2.5mm}
		\includegraphics[width=\linewidth]{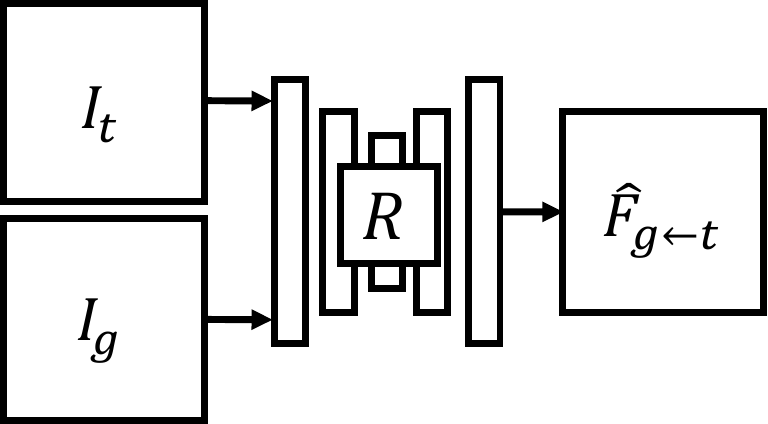}
		\caption{\small{Testing usage.}}
	\end{subfigure}
	\quad \quad
	\begin{subfigure}[b]{0.55\linewidth}
		\centering
		\includegraphics[width=\linewidth,trim={0 3mm 0 3mm},clip]{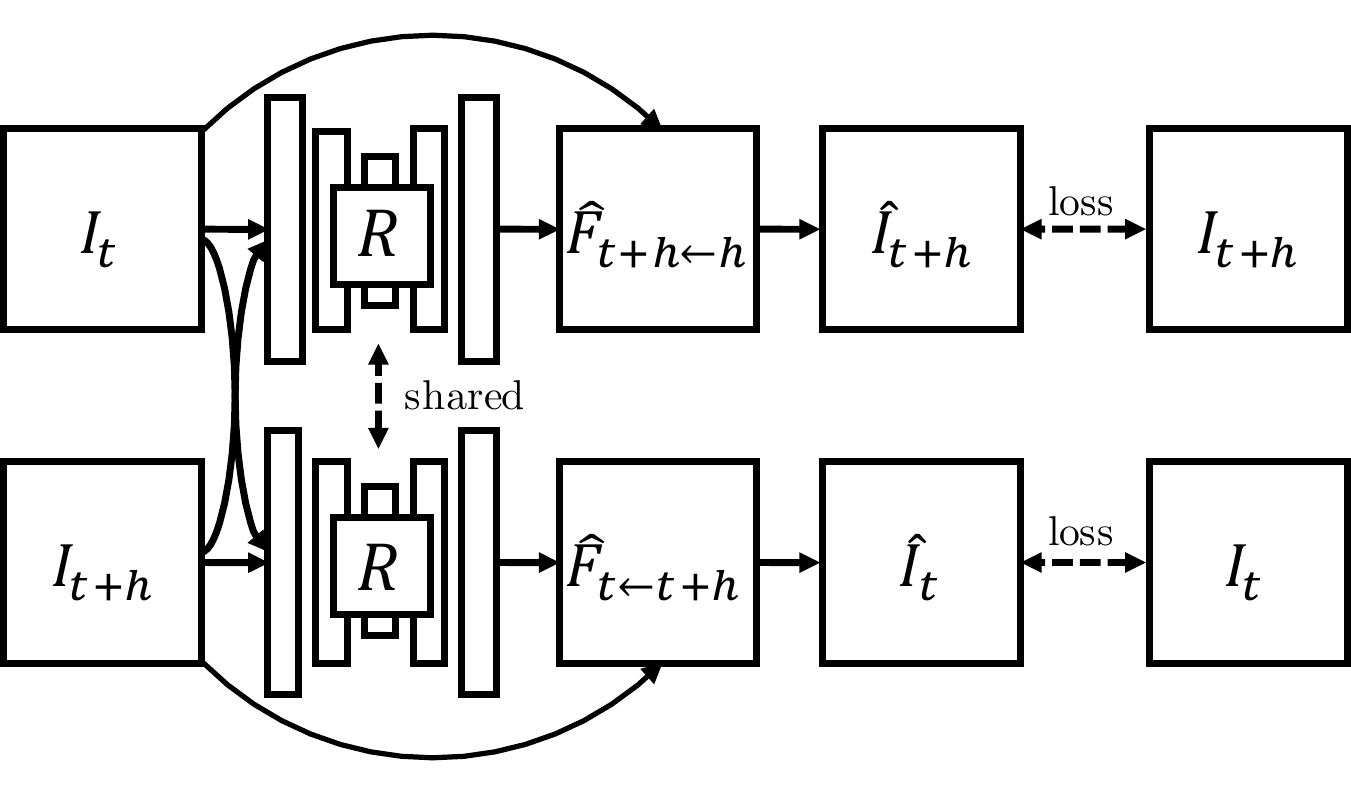}
		\caption{\small{Training usage.}}
		\label{fig:discrete}
	\end{subfigure}
	\vspace{-1mm}
	\caption{\small{(a) At test time the registration network registers the current image $I_t$ to the start image $I_0$ (top) and goal image $I_g$ (bottom), inferring the flow-fields $\hat{F}_{0 \leftarrow t}$ and $\hat{F}_{g \leftarrow t}$. (b) The registration network is trained by warping images from randomly selected timesteps along a trajectory to each other.
	}}
	\label{fig:registration_arch}
\end{figure}

\noindent \textbf{Test time procedure.}
We will first describe the registration scheme at test time (see Figure~\ref{fig:registration_arch}(a)). We separately register the current image $I_t$ to the start image $I_0$ and to the goal image $I_g$ by passing it into the registration network $R$, implemented as a fully-convolutional neural network. The registration network produces a flow map $\hat{F}_{0 \leftarrow t} \in \mathbb{R}^{H \times W \times 2}$, a vector field with the same size as the image, that describes the relative motion for every pixel between the two frames.
\begin{align}
\hat{F}_{0 \leftarrow t} = R(I_t, I_0) &&
\hat{F}_{g \leftarrow t} = R(I_t, I_g)
\end{align}
The flow map $\hat{F}_{0 \leftarrow t}$ can be used to warp the image of the current time step $t$ to the start image $I_0$, and $\hat{F}_{g \leftarrow t}$ can be used to warp from $I_t$ to $I_g$ (see Figure \ref{fig:reg_single} for an illustration). There is no difference to the warping operation used in the video prediction model, explained in section \ref{sec:model}, equation \ref{simple_dna}:
\begin{align}
\hat{I}_0 = \hat{F}_{0 \leftarrow t} \diamond  I_t &&
\hat{I}_g = \hat{F}_{g \leftarrow t} \diamond  I_t 
\end{align}
In essence for a current image $\hat{F}_{0 \leftarrow t}$ puts $I_t$ in correspondence with $I_0$, and $\hat{F}_{g \leftarrow t}$ puts $I_t$ in correspondence with $I_g$. The motivation for registering to both $I_0$ and $I_g$ is to increase accuracy and robustness. In principle, registering to either $I_0$ or $I_g$ is sufficient.
While the registration network is trained to perform a global registration between the images, we only evaluate it at the points $d_0$ and $d_g$ chosen by the user. This results in a cost function that ignores distractors. The flow map produced by the registration network is used to find the pixel locations corresponding to $d_0$ and $d_g$ in the current frame: 
\begin{align}
\hat{d}_{0,t} = d_0 + \hat{F}_{0 \leftarrow t}(d_0) &&
\hat{d}_{g,t} = d_g + \hat{F}_{g \leftarrow t}(d_g)
\label{eqn:warped_pos}
\end{align}

\begin{figure*}
	\centering
	\vspace{-0.1in}	\includegraphics[width=1\linewidth]{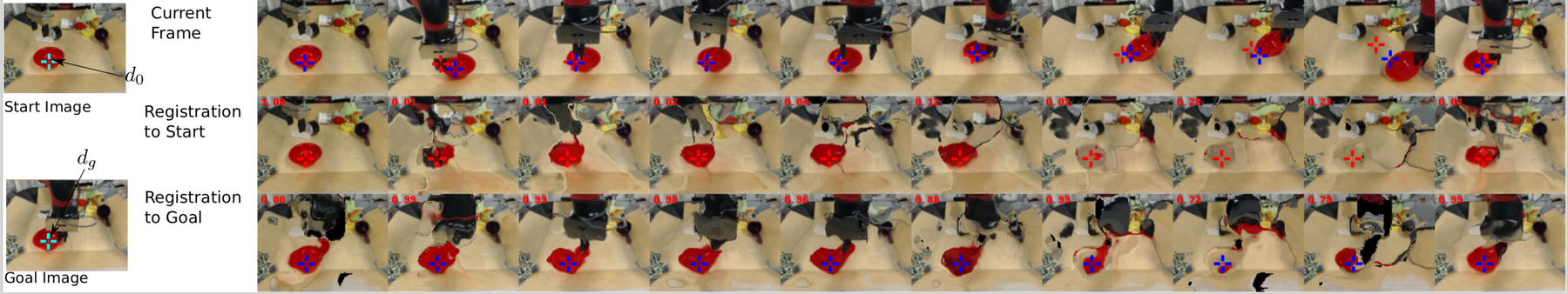}
	\caption{\small{Outputs of registration network. The first row shows the timesteps from left to right of a robot picking and moving a red bowl, the second row shows each image warped to the initial image via registration, and the third row shows the same for the goal image. A successful registration in this visualization would result in images that closely resemble the start- or goal image. In the first row, the locations where the designated pixel of the start image $d_0$ and the goal image $d_g$ are found are marked with red and blue crosses, respectively. It can be seen that the registration to the start image (red cross) is failing in the second to last time step, while the registration to the goal image (blue cross) succeeds for all time steps. The numbers in red, in the upper left corners indicate the trade off factors $\lambda$ between the views and are used as weighting factors for the planning cost. (Best viewed in PDF)}}
	\label{fig:tracking_overtime}
	\vspace{-0.2in}
\end{figure*}

For simplicity, we describe the case with a single designated pixel. In practice, instead of a single flow vector $\hat{F}_{0 \leftarrow t}(d_0)$ and $\hat{F}_{g \leftarrow t}(d_g)$, we consider a neighborhood of flow-vectors around $d_0$ and $d_g$ and take the median in the $x$ and $y$ directions, making the registration more stable.
\autoref{fig:tracking_overtime} visualizes an example tracking result while the gripper is moving an object.

\noindent \textbf{Registration-based pixel distance cost.}
Registration can fail when distances between objects in the images are large. During a motion, the registration to the first image typically becomes harder, while the registration to the goal image becomes easier. We propose a mechanism that estimates which image is registered correctly, allowing us to utilize only the successful registration for evaluating the planning cost. This mechanism gives a high weight $\lambda_i$ to pixel distance costs $c_i$ associated with a designated pixel $\hat{d}_{i,t}$ that is tracked successfully and a low, ideally zero, weight to a designated pixel where the registration is poor. We use the photometric distance between the true frame and the warped frame evaluated at $d_{0,i}$ and $d_{g,i}$ as an estimate for \emph{local} registration success. A low photometric error indicates that the registration network predicted a flow vector leading to a pixel with a similar color, thus indicating warping success. However this does not necessarily mean that the flow vector points to the correct location. For example, there could be several objects with the same color and the network could simply point to the wrong object. Letting $I_i(d_i)$ denote the pixel value in image $I_i$ for position $d_i$, and $\hat{I}_i(d_i)$ denote the corresponding pixel in the image warped by the registration function, we can define the general weighting factors $\lambda_i$ as:
\begin{align}
\lambda_i =  \frac{||I_i(d_i) - \hat{I_i}(d_i)||_2^{-1}}{\sum^N_j ||I_j(d_j) - \hat{I}_j(d_j)||^{-1}_2}.
\label{eqn:cost_avg}
\end{align}
where $\hat{I}_i = \hat{F}_{i \leftarrow t} \diamond I_t$. The MPC cost is computed as the average of the costs $c_i$ weighted by $\lambda_i$, where each $c_i$ is the expected distance (see equation \ref{eq:cost}) between the registered point $\hat{d}_{i,t}$ and the goal point $d_{g,i}$. Hence, the cost used for planning is $c = \sum_i \lambda_i c_i$.  In the case of the single view model and a single designated pixel, the index $i$ iterates over the start and goal image (and $N=2$).

The proposed weighting scheme can also be used with multiple designated pixels, as used in multi-task settings and multi-view models, which are explained in section \ref{sec:multiview}. The index $i$ then also loops over the views and indices of the designated pixels.

\noindent \textbf{Training procedure.}
The registration network is trained on the same data as the video prediction model, but it does not share parameters with it.\footnote{In principle, sharing parameters with the video prediction model might be beneficial, but this is left for future work.} Our approach is similar to the optic flow method proposed by \cite{meister2017unflow}. However, unlike this prior work, our method computes registrations for frames that might be many time steps apart, and the goal is not to extract optic flow, but rather to determine correspondences between potentially distant images. For training, two images are sampled at random times steps $t$ and $t+h$ along the trajectory and the images are warped to each other in both directions. 
\begin{align}
\hat{I}_{t} = \hat{F}_{t \leftarrow t +h} \diamond  I_{t+h} &&
\hat{I}_{t+h} = \hat{F}_{t+h \leftarrow t} \diamond  I_{t}
\end{align}
The network, which outputs $\hat{F}_{t \leftarrow t +h}$ and $\hat{F}_{t+h \leftarrow t}$, see Figure~\ref{fig:registration_arch} (b), is trained to minimize the photometric distance between $\hat{I}_t$ and $I_t$ and $\hat{I}_{t+h}$ and $I_{t+h}$, in addition to a smoothness regularizer that penalizes abrupt changes in the outputted flow-field. The details of this loss function follow prior work \cite{meister2017unflow}. We found that gradually increasing the temporal distance $h$ between the images during training yielded better final accuracy, as it creates a learning curriculum. The temporal distance is linearly increased from 1 step to 8 steps at 20k SGD steps. In total 60k iterations were taken.

The network $R$ is implemented as a fully convolutional network taking in two images stacked along the channel dimension. First the inputs are passed into three convolutional layers each followed by a bilinear downsampling operation. This is passed into three layers of convolution each followed by a bilinear upsampling operation (all convolutions use stride 1). By using bilinear sampling for increasing or decreasing image sizes we avoid artifacts that are caused by strided convolutions and deconvolutions.

\subsection{Classifier-Based Cost Functions}
\label{subsec:class_cost}
An alternative way to define the cost function is with a goal classifier. This type of cost function is particularly well-suited for tasks that can be completed in multiple ways. For example, for a task of rearranging a pair objects into relative positions, i.e. pushing the first object to the left of the second object, the absolute positions of the objects do not matter nor does the arm position. A classifier-based cost function allows the planner to discover any of the possible goal states. 

Unfortunately, a typical image classifier will require a large amount of labeled examples to learn, and we do not want to collect large datasets for each and every task. Instead, we aim to learn a goal classifier from only a few positive examples, using a meta-learning approach. A few positive examples of success are easy for people to provide and are the minimal information needed to convey a goal.

\newcommand{\task}{\mathcal{T}}
\newcommand{\data}{\mathcal{D}}
\newcommand{\obs}{\mathbf{o}}
\newcommand{\out}{y}
\newcommand{\posdata}{\data^+}
\newcommand{\testdata}{\data^\text{test}}
\newcommand{\loss}{\mathcal{L}}

Formally, we consider a goal classifier $\hat{\out} = f(\obs)$, where $\obs$ denotes the image observation, and $\hat{\out} \in [0,1]$ indicates the predicted probability of the observation being of a successful outcome of the task. Our objective is to infer a classifier for a new task $\task_j$ from a few positive examples of success, which are easy for a user to provide and encode the minimal information needed to convey a task. In other words, given a dataset $\posdata_j$ of $K$ examples of successful end states for a new task $\task_j$: $\data_j:=\{(\obs_k, 1) | k = 1...K\}_j$, our goal is to infer a classifier for task $\task_j$. 

\begin{figure}
    \centering
    \includegraphics[width=1.0\columnwidth]{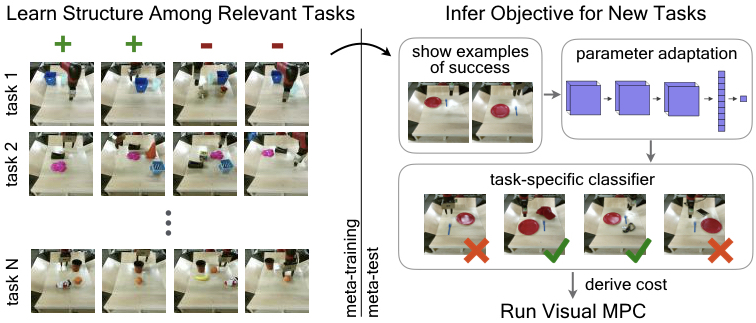}
    \caption{\small We propose a framework for quickly specifying visual goals. Our goal classifier is meta-trained with positive and negative examples for diverse tasks (left), which allows it to meta-learn that some factors matter for goals (e.g., relative positions of objects), while some do not (e.g. position of the arm). At meta-test time, this classifier can learn goals for new tasks from a few of examples of success (right - the goal is to place the fork to the right of the plate). The cost can be derived from the learned goal classifier for use with visual MPC.}
    \label{fig:cls_fig}
    \vspace{-0.3cm}
\end{figure}

\noindent \textbf{Meta-learning for few-shot goal inference.}
To solve the above problem, we propose learning a few-shot classifier that can infer the goal of a new task from a small set of goal examples, allowing the user to define a task from a few examples of success. To train the few-shot classifier, we first collect a dataset of both positive and negative examples for a wide range of tasks. We then use this data to learn how to learn goal classifiers from a few positive examples.
Our approach is illustrated in Figure~\ref{fig:cls_fig}.

We build upon model-agnostic meta-learning (MAML)~\cite{maml}, which learns initial parameters $\theta$ for model $f$ that can efficiently adapt to a new task with one or a few steps of gradient descent. Grant et al. \cite{caml} proposed an extension of MAML, referred to as concept acquisition through meta-learning (CAML), for learning to learn new concepts from positive examples alone. We apply CAML to the setting of acquiring goal classifiers from positive examples, using a meta-training data with both positive and negative examples. The result of the meta-training procedure is an initial set of parameters that can be used to learn new goal classifiers at test time.

\noindent \textbf{Test time procedure.}
At test time, the user provides a dataset $\posdata_j$ of $K$ examples of successful end states for a new task $\task_j$: $\data_j:=\{(\obs_k, 1) | k = 1...K\}_j$, which are then used to infer a task-specific goal classifier $C_j$. In particular, the meta-learned parameters $\theta$ are updated through gradient descent to adapt to task $\task_j$:

$$
C_j(\obs)
= f(\obs; \theta_j')
= f\big(\obs; \theta-\alpha \nabla_\theta \!\!\! \sum_{(\obs_n, \out_n)\in \posdata_j} \loss (\out_n, f(\obs_n; \theta)\big)
$$

where $\loss$ is the cross-entropy loss function, $\alpha$ is the step size, and $\theta'$ denotes the parameters updated through gradient descent on task $\task_j$.

During planning, the learned classifier $C_j$ takes as input an image generated by the video prediction model and outputs the predicted probability of the goal being achieved for the task specified by the few examples of success. To convert this into a cost function, we treat the probability of success as the planning cost for that observation. To reduce the effect of false positives and mis-calibrated predictions, we use the classifier conservatively by thresholding the predictions so that reward is only given for confident successes. Below this threshold, we give a reward of 0 and above this threshold, we provide the predicted probability as the reward.

\noindent \textbf{Training time procedure.}
During meta-training, we explicitly train for the ability to infer goal classifiers for the set of training tasks, $\{ \task_i \}$. We assume a small dataset $\data_i$ for each task $\task_i$, consisting of both positive and negative examples: $\data_i:= \{(\obs_n,\out_n) | n=1...N \}_i$. To learn the initial parameters $\theta$, we optimize the following objective using Adam~\cite{ADAM}:

$$
\min_\theta \sum_i \sum_{(\obs_n, y_n) \in \testdata_i} \loss(\out_n, f(\obs_n; \theta_i')) 
$$

In our experiments, our classifier is represented by a convolutional neural network, consisting of three convolutional layers, each followed by layer normalization and a ReLU non-linearity. After the final convolutional layer, a spatial soft-argmax operation extracts spatial feature points, which are then passed through fully-connected layers.

\subsection{When to Use Which Cost Function?}
\label{subsec:cost_discuission}

We have introduced three different forms of cost function, pixel distance based cost functions with and without registration, as well as classifier-based cost functions. Here we discuss the relative strengths and weaknesses of each.

Pixel distance based cost functions have the advantage that they allow moving objects precisely to target locations. They are also easy to specify, without requiring any example goal images, and therefore provide an easy and fast user interface. The pixel distance based cost function also has a high degree of robustness against distractor objects and clutter, since the optimizer can ignore the values of other pixels; this is important when targeting diverse real-world environments.
By incorporating an image of the goal, we can also add a registration mechanism to allow for more robust closed-loop control, at the cost of a more significant burden on the user.

The classifier-based cost function allows for solving more abstract tasks since it can capture invariances, such as the position of the arm, and settings where the absolute positions of an object is not relevant, such as positioning a cup in front of a plate, irrespective of where the plate is.  Providing a few example images takes more effort than specifying pixel locations but allows a broader range of goal sets to be specified.

\section{Trajectory Optimizer}

\begin{algorithm}[ht]
\caption{Planning in Visual MPC}
\label{alg:opt}
\begin{algorithmic}[1]
\State \textbf{Inputs:} Predictive model $g$, planning cost function $c$
\For{$t~=~0...T-1$}

\For{$i~=~0...n_{iter}-1$}
\If{$i==0$}
\State \begin{varwidth}[t]{\linewidth}
	Sample $M$ action sequences $\{a^{(m)}_{t:t+H-1}\}$ from \par $\mathcal N(0, I)$ or
	custom sampling distribution
\end{varwidth}
\Else
\State \begin{varwidth}[t]{\linewidth}
	Sample $M$ action sequences ${a^{(m)}_{t:t+H-1}}$ from \par 
	$\mathcal N(\mu^{(i)}, \Sigma^{(i)})$
\end{varwidth}
\EndIf
\State 
\begin{varwidth}[t]{\linewidth}
Check if sampled actions are within \par
admissible range, otherwise resample.
\end{varwidth}
\State  \begin{varwidth}[t]{\linewidth}
	Use $g$ to predict future  image sequences $\hat{I}_{t:t+H-1}^{(m)}$\\ and probability distributions $\hat{P}_{t:t+H-1}^{(m)}$
\end{varwidth}
\State Evaluate action sequences using a cost function $c$
\State  \begin{varwidth}[t]{\linewidth}
	Fit a diagonal Gaussian to the $k$  action samples\\ with lowest cost,
	yielding $\mu^{(i)}, \Sigma^{(i)}$
\end{varwidth}
\EndFor
\State Apply first action of best action sequence to robot
\EndFor
\end{algorithmic}
\end{algorithm}

\label{sec:optimizer}
The role of the optimizer is to find actions sequences $a_{1:T}$ that minimize the sum of the costs $c_{1:T}$ along the planning horizon $T$. We use a simple stochastic optimization procedure for this, based on the cross-entropy method (CEM), a gradient-free optimization procedure. CEM consists of iteratively resampling action sequences and refitting Gaussian distributions to the actions with the best predicted cost.

Although a variety of trajectory optimization methods may be suitable, one advantage of the stochastic optimization procedure is that it allows us to easily ensure that actions stay within the distribution of actions the model encountered during training. This is crucial to ensure that the model does not receive out-of-distribution inputs and makes valid predictions.  Algorithm \ref{alg:opt} illustrates the planning process. In practice this can be achieved by defining admissible ranges for each dimension of the action vector and rejecting a sample if it is outside of the admissible range. 

In the appendix \ref{sec:cem_improv} we present a few improvements to the CEM optimizer for visual MPC.
\section{Custom Action Sampling Distributions}

\label{sec:system}
When collecting data by sampling from simple distributions, such as a multivariate Gaussian, the skills that emerged were found to be generally restricted to pushing and dragging objects. This is because with simple distributions, it is very unlikely to visit states like picking up and placing of objects or folding cloth. Not only would the model be imprecise for these kinds of states, but also during planning it would be unlikely to \emph{find} action sequences that grasp an object or fold an item of clothing.
We therefore explore how the sampling distribution used both in data collection and sampling-based planning can be changed to visit these, otherwise unlikely, states more frequently, allowing more complex behavior to emerge. 

To allow picking up and placing of objects as well as folding of cloth to occur more frequently, we incorporate a simple ``reflex'' during data collection, where the gripper automatically closes, when the height of the wrist above the table is lower than a small threshold. This reflex is inspired by the palmar reflex observed in infants~\cite{grasping_fetal}. With this primitive, when collecting data with rigid objects about 20\% of trajectories included some sort of grasp. For deformable objects such as towels and cloth, this primitive helps increasing the likelihood of encountering states where cloths are folded. We found that the primitive can be slightly adapted to avoid cloths becoming tangled up. More details are provided in Appendix \ref{sec:folding_sampling}.

It is worth noting that, other than this reflex, no grasping-specific or folding-specific engineering was applied to the policy, allowing a joint pushing, grasping and folding policy to emerge through planning (see figure \ref{fig:push_grasp} in the appendix). In our experiments, we evaluate our method using data obtained both with and without the grasping reflex, evaluating both purely non-prehensile and combined prehensile and non-prehensile manipulation.

\section{Multi-View Visual MPC}
\label{sec:multiview}
\begin{wrapfigure}{r}{.5\columnwidth}
	\centering
	\includegraphics[width=0.5\columnwidth]{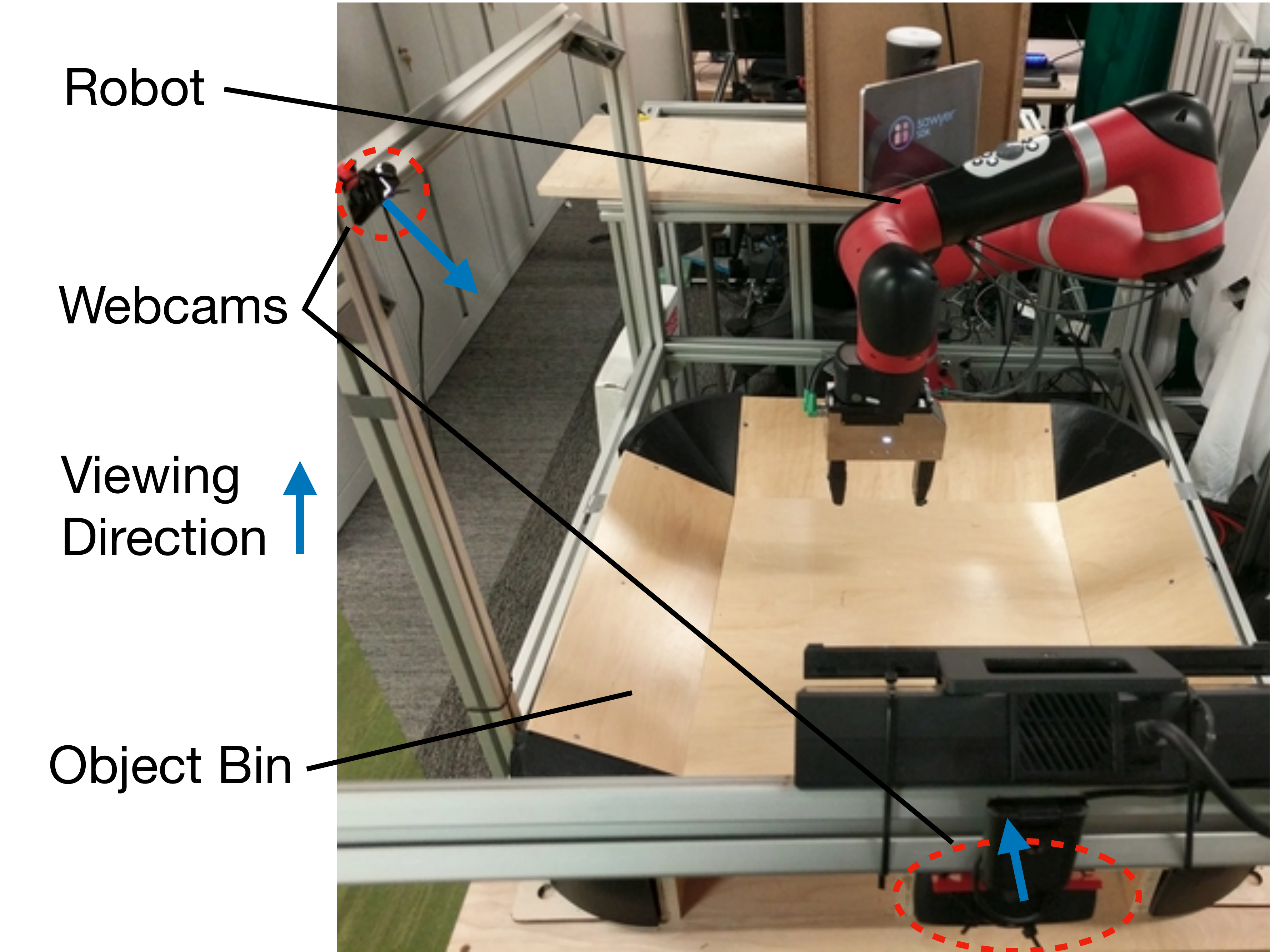}
	\caption{\small{Robot setup, with 2 standard web-cams arranged at different viewing angles.}}	\label{fig:robot_setup}
\end{wrapfigure}

The visual MPC algorithm as described so far is only able to solve manipulation tasks specified in 2D, like rearranging objects on the table. However, this can impose severe limitations; for example, a task such as lifting an object to a particular position in 3D cannot be fully specified with a single view, since it would be ambiguous. 
We use a combination of two views, taken from two cameras arranged appropriately, to jointly define a 3D task. Figure \ref{fig:robot_setup} shows the robot setup, including two standard webcams observing the workspace from different angles. The registration method described in the previous section is used separately per view to allow for dynamic retrying and solving temporally extended tasks. The planning costs from each view are combined using weighted averaging where the weights are provided by the registration network (see equation \ref{eqn:cost_avg}). Rows 5 and 6 of figure \ref{fig:tile_2} show a 3D object positioning task, where an object needs to be positioned at a particular point in 3D space. This task needs two views to be fully specified.

\section{Experimental Evaluation}
\label{sec:experiments}
In this section we present both qualitative and quantitative  performance evaluations of visual MPC on various manipulation tasks assessing the degree of generalization and comparing different prediction models and cost functions and with a hand-crafted baseline.
In Figures \ref{fig:example_traj} and \ref{fig:tile_2} we present a set of qualitative experiments showing that visual MPC trained fully self-supervised is capable of solving a wide range of complex tasks.
Videos for the qualitative examples are at the following webpage\footnote{Videos \& code: \url{https://sites.google.com/view/visualforesight/}}.
In order to perform quantitative comparisons, we define a set of tasks where the robot is required to move object(s) into a goal configuration. For measuring success, we use a distance-based evaluation where a human annotates the positions of the objects after pushing allowing us to compute the remaining distance to the goal.

\subsection{Comparing Video Prediction Architectures}
\label{subsec:sna_experiments}
We first aim to answer the question: Does  visual MPC using the occlusion-aware SNA video prediction model that includes temporal skip connections outperform visual MPC with the dynamic neural advection model (DNA)\cite{foresight} \emph{without} temporal skip-connections?

\begin{table}
\centering
{\footnotesize
\begin{tabular}{lcc}
	\toprule
         &  \thead{moved imp. \\ $\pm$ std err. of mean} &   \thead{stationary imp. \\ $\pm$ std err. of mean}  \\
         \midrule
  DNA \cite{foresight} & 0.83 $\pm$0.25 &  -1.1 $\pm$ 0.2\\ 
  SNA & \textbf{10.6 $\pm$ 0.82} & \textbf{-1.5 $\pm$ 0.2} \\
  \bottomrule
\end{tabular}
}
\caption{Results for multi-objective pushing on 8 object/goal configurations with 2 seen and 2 novel objects. Values indicate improvement in distance from starting position, higher is better. Units are pixels in the 64x64 images.} 
\label{table:mult_obj}
\end{table}
To examine whether our skip-connection model (SNA) helps with handling occlusions, we devised a task that requires the robot to push one object, while keeping another object stationary. When the stationary object is in the way, the robot must move the target object around it. This is illustrated on the left side of \autoref{fig:goingaroundocclusion} in the appendix. While pushing the target object, the gripper may occlude the stationary object, and the task can only be performed successfully if the model can make accurate predictions through this occlusion. These tasks are specified by selecting one starting pixel on the target object, a goal pixel location for the target object, and commanding the obstacle to remain stationary by selecting the same pixel on the obstacle for both start and goal. 

We use four different object arrangements with two training objects and two objects that were not seen during training. We find that, in most cases, the SNA model is able to find a valid trajectory, while the DNA model, that is not able to handle occlusion, is mostly unable to find a solution. The results of our quantitative comparisons are shown in \autoref{table:mult_obj}, indicating that temporal skip-connections indeed help with handling occlusion in combined pushing and obstacle avoidance tasks. 

\subsection{Evaluating Registration-Based Cost Functions}
\label{susbsec:reg_cost_exp}

\begin{table}
	{\footnotesize
		\begin{center}
			\begin{tabular}{lcc}
				\toprule
				& Short & Long \\
				\midrule
				Visual MPC $+$ predictor propagation  & 83\% & 20\% \\
				Visual MPC $+$ OpenCV tracking  & 83\%  & 45\% \\
				Visual MPC $+$ registration network & 83\% & \textbf{66\%}  \\
				\bottomrule
			\end{tabular}
		\end{center}
	}
	\caption{\small Success rate for long-distance pushing experiment with 20 different object/goal configurations and short-distance experiment with 15 object/goal configurations. Success is defined as bringing the object closer than 15 pixels to the goal, which corresponds to around $7.5cm$.}
	\label{table:res_long_short}
\end{table}
In this section we ask: How important  is it to update the model's belief of where the target objects currently are? 
We first provide two qualitative examples: In example (5)-(6) of Figure \ref{fig:tile_2} the task is to bring the stuffed animal to a particular location in 3D-space on the other side of the arena. To test the system's reaction to perturbations that could be encountered in open-world settings, during execution a person knocks the object out of the robot's hand (in the 3rd frame). The experiment shows that visual MPC is able to naturally perform a new grasp attempt and bring the object to the goal. This trajectory is easier to view in the supplementary video.

In Figure \ref{fig:push_retry} in the appendix, the task is to push the bottle to the point marked with the green dot. In the beginning of the trajectory the object behaves differently than expected, it moves downwards instead of to the right. However the system recovers from the initial failure and still pushes the object to the goal.

The next question we investigate is: How much does tracking the target object using the learned registration matter for short horizon versus long horizon tasks? 
In this experiment, we disable the gripper control, which requires the robot to push objects to the target. We compare two variants of updating the positions of the designated pixel when using a pixel-distance based cost function. The first is a cost function that uses our registration-based method, trained in a fully self-supervised fashion, and the second is with a cost function that uses off-the shelf tracking from OpenCV \cite{babenko2009visual}. Additionally we compare to visual MPC,
which uses the video-prediction model's own prior predictions to update the current position of the designated pixel, rather than tracking the object with registration or tracking.

We evaluate our method on 20 long-distance and 15 short-distance pushing tasks. For long distance tasks the initial distance between the object and its goal position is $30cm$ while for short distance tasks it is $15cm$. Table \ref{table:res_long_short} lists quantitative comparisons showing that on the long distance experiment
visual MPC using the registration-based cost not only outperforms prior work \cite{sna}, but also outperforms the hand-designed, supervised object tracker \cite{babenko2009visual}. By contrast, for the short distance experiment,
all methods perform comparably. Thus, theses results demonstrate the importance of tracking the position of the target object for \emph{long-horizon tasks}, while for short-horizon tasks object tracking appears to be irrelevant. 
\subsection{Evaluating Classifier-Based Cost Function}
\label{subsec:eval_classifier}

\begin{figure}
	\centering
	\includegraphics[width=1.0\columnwidth]{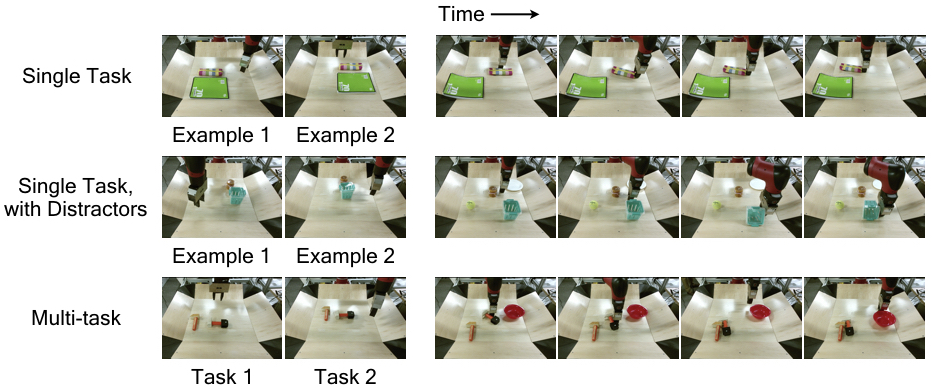}
	\caption{\small Object arrangement performance of our goal classifier with distractor objects and with two tasks. The left shows a subset of the 5 positive examples that are provided for inferring the goal classifier(s), while the right shows the robot executing the specified task(s) via visual planning.}
	\label{fig:cls_results}
	\vspace{-0.3cm}
\end{figure}

The goal of the classifier-based cost function is to provide an easy way to compute an objective for new tasks from a few observations of success for that task, so we compare our approach to alternative and prior methods for doing so under the same assumptions: pixel distance and latent space distance. In the latter, we measure the distance between the current and goal observations in a learned latent space, obtained by training an autoencoder (DSAE) \cite{dsae} on the same data used for our classifier. Since we are considering a different form of task specification incompatible with user-specified pixels, we do not compare the classifier-based cost function to the cost function based on designated pixels.

To collect data for meta-training the classifier, we randomly select a pair of objects from our set of training objects, and position them in many different relative positions, recording the image for each configuration. Each task corresponds to a particular relative positioning of two objects, e.g. the first object to the left of the second, and we construct positive and negative examples for each task by labeling the aforementioned images. We randomly position the arm in each image, as it is not a determiner of task success. A good classifier should ignore the position of the arm. We also include randomly-positioned distractor objects in about a third of the collected images.

We evaluate the classifier-based cost function in three different experimental settings. In the first setting, the goal is to arrange two objects into a specified relative arrangement. The second setting is the same, but with distractor objects present. In the final and most challenging setting, the goal is to achieve two tasks in sequence. We provide positive examples for both tasks, infer the classifier for both, perform MPC for the first task until completion, followed by MPC for the second task. The arrangements of the evaluation tasks were chosen among the eight principal directions (N, NE, E, SE, etc.). To evaluate the ability to generalize to new goals and settings, we use novel, held-out objects for all of the task and distractor objects in our evaluation.

\begin{figure}
    \centering
    \includegraphics[width=0.48\textwidth]{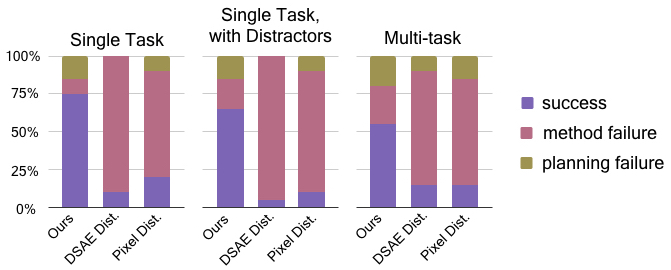}
    \caption{\small Quantitative performance of visual planning for object rearrangement tasks across different goal specification methods: our meta-learned classifier, DSAE~\cite{dsae}, and pixel error. Where possible, we include break down the cause of failures into errors caused by inaccurate prediction or planning and those caused by an inaccurate goal classifier.}
    \label{fig:cls_charts}
    \vspace{-0.3cm}
\end{figure}

We qualitatively visualize the tasks in Figure~\ref{fig:cls_results}. On the left, we show a subset of the five images provided to illustrate the task(s), and on the left, we show the motions performed by the robot. We see that the robot is able to execute motions which lead to a correct relative positioning of the objects.
We quantitatively evaluate the three cost functions across 20 tasks, including $10$ unique object pairs. A task was considered successfully completed if more than half of the object was correctly positioned relative to the other. The results, shown in Figure~\ref{fig:cls_charts}, indicate that the distance-based metrics struggle to infer the goal of the task, while our approach leads to substantially more successful behavior on average.

\subsection{Evaluating Multi-Task Performance}
\label{subsec:multi_task_bench}
\begin{figure*}
	\centering
	\includegraphics[width=2.05\columnwidth]{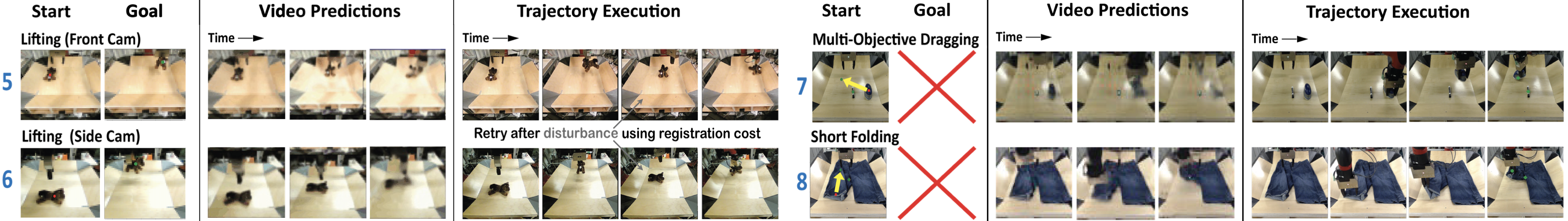}
	\caption{Visual MPC successfully solves a wide variety of tasks including multi-objective tasks, such as placing an object on a plate (row 5 and 6), object positioning with obstacle avoidance (row 7) and folding shorts (row 8). Zoom in on PDF.   
		\label{fig:tile_2}}
\end{figure*}

One of the key motivations for visual MPC is to build a system that can solve a \emph{wide variety} of different tasks, involving completely different objects, physics and, objectives.
Examples for tasks that can be solved with visual MPC are shown in Figure \ref{fig:example_traj} and \ref{fig:tile_2}. Task 1 in Figure~\ref{fig:example_traj} shows a ``placing task" where an object needs to be grasped and placed onto a plate while not displacing the plate. Task 2 is an object rearrangement tasks. The example shown in Task 4 and all examples in Figure \ref{fig:cls_results} show relative object rearrangement tasks. Examples 5 and 6 show the same 3D object positioning tasks from different views. In Task 7, the goal is to move the black object to the goal location while avoiding the obstacle in the middle which is marked with a designated- and goal pixel. We also demonstrate that visual MPC -- without modifications to the algorithm -- solves tasks involving deformable objects such as a task where a towel needs to be wrapped around an object (Task 3), or folding a pair of shorts (Task 8). To the best of our knowledge this is the first algorithm for robotic manipulation handling both rigid and deformable objects.
For a full illustration of each of these tasks, we encourage the reader to watch the supplementary video.

The generality of visual MPC mainly stems from two components --- the generality of the visual dynamics model and the generality of the task definition.
We found that the dynamics model often generalizes well to objects outside of the training set, if they have similar properties to the objects it was trained with. For example, Task 8 in Figure \ref{fig:tile_2} shows the model predicting a pair of shorts being folded. We observed that a model, which was only provided videos of towels during training, generalized to shorts, although it had never seen them before. In all of the qualitative examples, the predictions are performed by the \emph{same model}.
We found that the model sometimes exhibits confusion about whether an object follows the dynamics of a cloth or rigid objects, which is likely caused by a lack of training data in the particular regime. To overcome this issue we add a binary token to the state vector indicating whether the object in the bin is hard or soft. 
We expect that adding more training data would remove the need for this indicator and allow the model to infer material properties directly from images.

The ability to specify tasks in multiple different ways adds to the flexibility of the proposed system. Using designated pixels, object positioning tasks can be defined in 3D space, as shown in Task 1 and 2 in Figure~\ref{fig:example_traj} and task 5-6 in Figure~\ref{fig:tile_2}. When adding a goal image, the positioning accuracy can be improved by utilizing the registration scheme discussed in Section \ref{subsec:reg_cost}.
For tasks where we care about \emph{relative} rather than absolute positioning, a meta-learned classifier can be used, as discussed in Section \ref{subsec:class_cost}.

Next, we present a quantitative evaluation to answer the following question: How does visual MPC compare to a hand-engineered baseline on a large number of diverse tasks? For this comparison, we engineered a simple trajectory generator to perform a grasp at the location of the initial designated pixel, lift the arm, and bring it to the position of the goal pixel. Camera calibration was performed to carry out the necessary conversions between image-space and robot work-space coordinates, which was not required for our visual MPC method. For simplicity, the baseline controller executes in open loop. Therefore, to allow for a fair comparison, visual MPC is also executed open-loop, i.e. no registration or tracking is used.
Altogether we selected 16 tasks, including the qualitative examples presented earlier.
The quantitative comparison is shown in Table \ref{table:cloth_folding}, illustrating that visual MPC substantially outperforms this baseline.
Visual MPC succeeded for most of the tasks. While the baseline succeeded for some of the cloth folding tasks, it failed for almost all of the object relocation tasks. This indicates that an implicit understanding of physics, as captured by our video prediction models, is indeed essential for performing this diverse range of object relocation and manipulation tasks, and the model must perform non-trivial physical reasoning beyond simply placing and moving the end-effector.

\label{subsec:cloth_folding_data}
\begin{table}
\centering
{\footnotesize
\begin{tabular}{lcc}
	\toprule
         &  \thead{\% of Trials with \\ Final Pixel Distance $< 15$}   \\
         \midrule
  Visual MPC & \textbf{75\%} \\ 
  Calibrated Camera Baseline & 18.75 \% \\
  \bottomrule
\end{tabular}
}
\caption{Results for a multi-task experiment of 10 hard object pushing and grasping tasks, along with 6 cloth folding tasks, evaluating using a single model. Values indicate the percentage of trials that ended with the object pixel closer than 15 pixels to the designated goal. Higher is better.} 
\label{table:cloth_folding}
\end{table}

\subsection{Discussion of Experimental Results}

Generalization to many distinct tasks in visually diverse settings is arguably one of the biggest challenges in reinforcement learning and robotics today. While deep learning has relieved us from much of the problem-specific engineering, most of the works either require extensive amounts of labeled data or focus on the mastery of single tasks while relying on human-provided reward signals. 
From the experiments with visual MPC, especially the qualitative examples and the multi-task experiment, we can conclude that visual MPC \emph{generalizes} to a wide range of tasks it has never seen during training. This is in contrast to many model-free approaches for robotic control which often struggle to perform well on novel tasks. Most of the generalization performance is likely a result of large-scale self-supervised learning, which allows to acquire a rich, task-agnostic dynamics model of the environment.  %

\section{Conclusion}
We presented an algorithm that leverages self-supervision from visual prediction to learn a deep dynamics model on images, and show how it can be embedded into a  planning framework to solve a variety of robotic control tasks. We demonstrate that visual model-predictive control is able to successfully perform multi-object manipulation, pushing, picking and placing, and cloth-folding tasks -- all within a single framework. 

\noindent \textbf{Limitations.}
The main limitations of the presented framework are that all target objects need to be visible throughout execution, it is currently not possible to handle partially observed domains. This is especially important for tasks that require objects to be brought into occlusion (or taken out of occlusion), for example putting an object in a box and closing it. Another limitation is that the tasks are still of only medium duration and usually only touch one or two objects. Longer-term planning remains an open problem. Lastly, the fidelity of object positioning is still significantly below what humans can achieve.

\noindent \textbf{Possible future directions.}
The key advantage of a model-based deep-reinforcement learning algorithm like visual MPC is that it \emph{generalizes} to tasks it has never encountered before. This makes visual MPC a good candidate for a building block of future robotic manipulation systems that will be able solve an even wider range of complex tasks with much longer horizons.

\ifCLASSOPTIONcompsoc
\else
\fi

\ifCLASSOPTIONcaptionsoff
  \newpage
\fi

\bibliographystyle{IEEEtran}
\bibliography{mybib}
\vskip -13mm

\begin{IEEEbiographynophoto}{Frederik Ebert}
Frederik Ebert received a BS in Mechatronics and Information Technology as well a MS in "Robotics, Cognition, Intelligence (RCI)" from the Technical University of Munich (TUM). He is currently a PhD student at Berkeley Artifical Intelligence Research (BAIR),  where he focuses on developing algorithms for robotic manipulation combining ideas from computer vision, machine learning, and control.
\end{IEEEbiographynophoto}
\vskip -13mm

\begin{IEEEbiographynophoto}{Chelsea Finn} is a research scientist at Google and a post-doctoral scholar at UC Berkeley, and will join the Computer Science faculty at Stanford in 2019. Her research focuses on algorithms that can enable agents to autonomously learn a range of complex skills. 
She received a BS in Electrical Engineering and Computer Science from MIT and a PhD in Computer Science from UC Berkeley.
\end{IEEEbiographynophoto}
\vskip -13mm

\begin{IEEEbiographynophoto}{Sudeep Dasari} is a 4th year student at UC Berkeley pursuing a B.S in Electrical Engineering and Computer Science. His primary research interests are computer vision, machine learning, and robotic control.
\end{IEEEbiographynophoto}
\vskip -13mm

\begin{IEEEbiographynophoto}{Annie Xie}
is pursuing a B.S. degree in Electrical Engineering and Computer Science at UC Berkeley. Her research interests are in the areas of computer vision and robot learning.
\end{IEEEbiographynophoto}
\vskip -13mm

\begin{IEEEbiographynophoto}{Alex Lee} received a BS in Electrical Engineering and Computer Science from UC Berkeley and is currently pursuing a PhD in Computer Science from UC Berkeley. His work focuses on algorithms that can enable robots to learn complex sensorimotor skills.
\end{IEEEbiographynophoto}
\vskip -13mm

\begin{IEEEbiographynophoto}{Sergey Levine} received a BS, MS, and PhD in Computer Science from Stanford. He is currently on the faculty of the Department of Electrical Engineering and Computer Sciences at UC Berkeley. His work focuses on machine learning for decision making and control, with an emphasis on deep learning and reinforcement learning algorithms.
\end{IEEEbiographynophoto}

\appendices
\newpage

\section{Skip Connection Neural Advection Model}
\label{sec:skipcon}

Our video prediction model, shown in ~\autoref{fig:prediction_model}, is inspired by the dynamic neural advection (DNA) model proposed by~\cite{finn_nips} and it is a variant of the SNA model proposed by~\cite{sna}. The model uses a convolutional LSTM~\cite{convlstm} to predict future frames. The prediction is initialized with an initial sequence of 2 ground truth frames, and predicts 13 futures frames. The model predicts these frames by iteratively making next-frame predictions and feeding those predictions back to itself. Each predicted frame, is given by a compositing layer, which composes intermediate frames with predicted compositing masks. The intermediate frames include the previous 2 frames and transformed versions them. To easily allow various transformations for different parts of the image, we predict 8 transformed frames, 4 of which are transformed from the previous frame, and the other 4 from the frame 2 steps in the past. These intermediate frames are then combined with compositing masks, which are also predicted by the convolutional LSTM. For simplicity, we collectively refer to these transformations as a single transformation $\hat{F}_{t+1 \leftarrow t}$ in~\autoref{simple_dna}. In addition, the first frame of the sequence is also given as one of the intermediate frames~\cite{sna}.

To enable action conditioning, the robot action at each time step is passed to all the convolutional layers of the main network, by concatenating it along the channel dimension of the inputs of these layers. Since they are vectors with no spatial dimensions, they are replicated spatially to match the spatial dimensions of the inputs.

The SNA variant that we use incorporate the architectural improvements proposed by~\cite{savp}. Each convolutional layer is followed by instance normalization \cite{instancenorm} and ReLU activations. We also use instance normalization on the LSTM pre-activations (i.e., the input, forget, and output gates, as well as the transformed and next cell of the LSTM). In addition, we modify the spatial downsampling and upsampling mechanisms. Standard subsampling and upsampling between convolutions is known to produce artifacts for dense image generation tasks~\cite{odena2016deconvolution,Niklaus_ICCV_2017}. In the encoding layers, we reduce the spatial resolution of the feature maps by average pooling, and in the decoding layers, we increase the resolution by using bilinear interpolation. All convolutions in the generator use a stride of 1. The actions are concatenated to the inputs of all the convolutional layers of the main network, as opposed to only the bottleneck.

\begin{wrapfigure}{r}{.37\columnwidth}
	\centering
	\includegraphics[width=0.37\columnwidth]{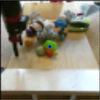}
	\caption{The blue dot indicates the designated pixel}
	\label{fig:desig_pix_bluedot}
\end{wrapfigure}

\begin{figure}
	\centering
	\begin{subfigure}{0.9\columnwidth}
		\centering
		\includegraphics[width=1.\linewidth]{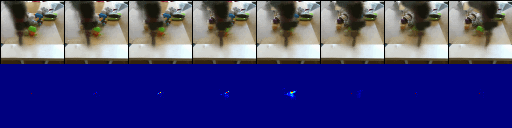}
		\caption{Skip connection neural advection (SNA) does not erase or move objects in the background}
		\label{fig:Ng1}
	\end{subfigure}
	\begin{subfigure}{0.9\columnwidth}
		\centering
		\includegraphics[width=1.0\linewidth]{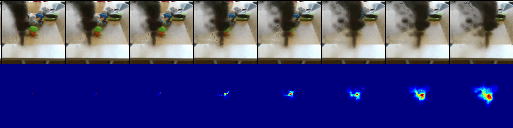}
		\caption{Standard DNA \cite{foresight} exhibits undesirable movement of the distribution $P_{d}(t)$ and erases the background}
	\end{subfigure}
	\caption{
		Top rows: Predicted images of arm moving \textit{in front of} green object with designated pixel (as indicated in \autoref{fig:desig_pix_bluedot}). 
		Bottom rows: Predicted probability distributions $P_{d}(t)$ of designated pixel obtained by repeatedly applying transformations.}
	\label{fig:pix_reappear}
\end{figure}

We provide an example of the skip connection neural advection (SNA) model recovering from occlusion in \autoref{fig:pix_reappear}. In this figure, the arm is predicted to move in front of the designated pixel, marked in blue in \autoref{fig:desig_pix_bluedot}. The predictions of the DNA model, shown in figure \autoref{fig:pix_reappear}(b), contain incorrect motion of the marked object, as shown in the heatmaps visualizing $\hat{P}_t$, although the arm actually passes in front of it. This is because the DNA model cannot recover information about an object that it has `overwritten' during its predictions, causing the model to predict that the pixel \emph{moves with the arm}. We identified this as one of the major causes of planning failure using the DNA model. By contrast our SNA model predicts that the occluded object will not move, shown in figure  \autoref{fig:pix_reappear}(a).

\section{Improved Action Sampling Distributions for Data Collection}
\label{sec:folding_sampling}
In order to collect meaningful interaction data for learning folding of deformable objects such as towels and cloth, we found that the grasping primitive can be slightly adapted to increase the likelihood of encountering states where cloths are actually folded. When using actions sampled from a simple distribution or the previously-described distribution, clothing would become tangled up. To improve the efficiency of folding cloths we use an action primitive similar to the grasping primitive, but additionally we reduce lateral motion of the end-effector when the gripper is close to the table, thus reducing events where cloths become tangled up.

\begin{figure*}
    \centering    \includegraphics[width=1.0\textwidth]{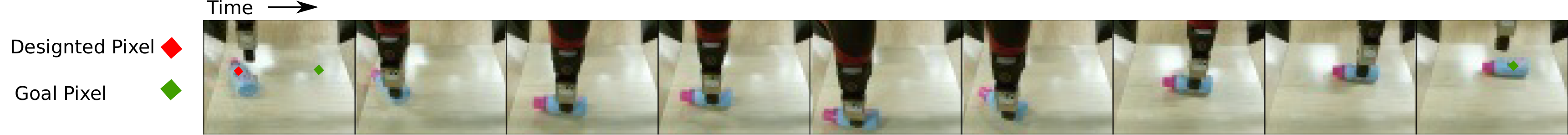}
    \caption{\small{Applying our method to a pushing task. In the first 3 time instants the object behaves unexpectedly, moving down. The tracking then allows the robot to retry, allowing it to eventually bring the object to the goal.}}
    \label{fig:push_retry}
\end{figure*}

\begin{figure*}
	\centering
	\includegraphics[width=1.0\textwidth]{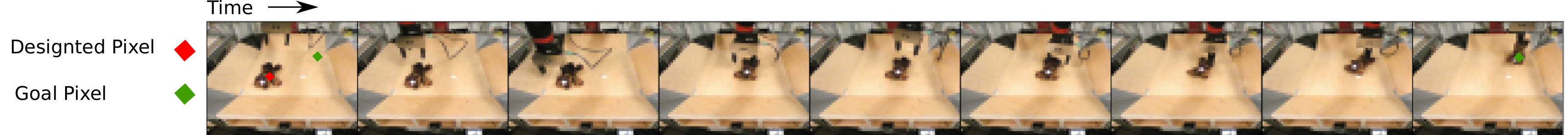}
	\caption{\small{Retrying behavior of our method combining prehensile and non-prehensile manipulation. In the first 4 time instants shown the robot pushes the object. It then loses the object, and decides to grasp it pulling it all the way to the goal. Retrying is enabled by applying the learned registration to both camera views (here we only show the front view).}}
	\label{fig:push_grasp}
\end{figure*}

\begin{figure*}
	\centering
	\includegraphics[width=1\linewidth]{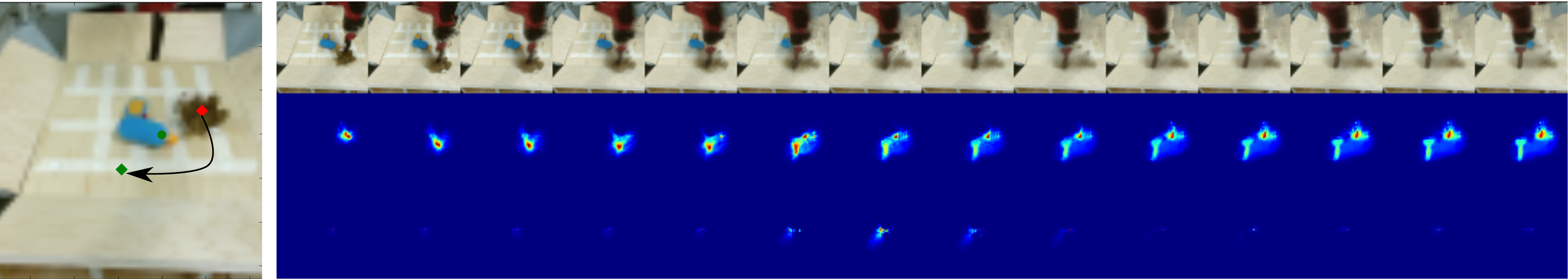}
	\caption{Left: Task setup with green dot marking the obstacle. Right, first row: the predicted frames generated by SNA. Second row: the probability distribution of the designated pixel on the \textit{moving} object (brown stuffed animal). Note that part of the distribution shifts down and left, which is the indicated goal. Third row: the probability distribution of the designated pixel on the obstacle-object (blue power drill). Although the distribution increases in entropy during the occlusion (in the middle), it then recovers and remains on its original position.
		\label{fig:goingaroundocclusion}}
\end{figure*}

\section{Improvements of the CEM-Optimizer}
\label{sec:cem_improv}
In the model-predictive control setting, the action sequences found by the optimizer can be very different between execution real-world times steps. For example at one time step the optimizer might find a pushing action leading towards the goal and in the next time step it determines a grasping action to be optimal to reach the goal. Na\"{i}ve replanning at every time step can then result in alternating between a pushing and a grasping attempt indefinitely causing the agent to get stuck and not making any progress towards to goal. 

We can resolve this problem by modifying the sampling distribution of the first iteration of CEM so that the optimizer commits to the plan found in the previous time step. In the simplest version of CEM the sampling distribution at first iteration of CEM is chosen to be a Gaussian with diagonal covariance matrix and zero mean. We instead use the best action sequence found in the optimization of the \emph{previous} real-world time step as the mean for sampling new actions in the \emph{current} real-world time-step. Since this action sequence is optimized for the previous time step we only use the values $a_{2:T}$ and omit the first action. To sample actions close to the action sequence from the previous time step we reduce the entries of the diagonal covariance matrix for the first $T-1$ time steps. It is crucial that the last entry of the covariance matrix at the end of the horizon is not reduced otherwise no exploration could happen for the last time step causing poor performance at later time steps.

\section{Experimental Setup}
\label{sec:experiment_setup}
To train both our video-prediction and registration models, we collected 20,000 trajectories of pushing motions and 15,000 trajectories with gripper control, where the robot randomly picks and moves objects using the grasping reflex described in section \ref{sec:system}. The data collection process is fully autonomous, requiring human intervention only to change out the objects in front of the robot. The action space consisted of relative movements of the end-effector in cartesian space along the $x$, $y$, and $z$ axes, and for some parts of the dataset we also added azimuthal rotations of the gripper and a grasping action.

\section{Experimental Evaluation}

\autoref{fig:goingaroundocclusion} shows an example of the SNA model successfully predicting the position of the obstacle through an occlusion and finding a trajectory that avoids the obstacle.

\section{Simulated Experiments}

In order to provide a more controlled comparison, we also set up a realistic simulation environment using MuJoCo \cite{todorov2012mujoco}, which includes a robotic manipulator controlled via Cartesian position control, similar to our real world setup, pushing randomly-generated L-shaped objects with random colors (see details in supplementary materials). 
We trained the same video prediction model in this environment, and set up 50 evaluation tasks where blocks must be pushed to target locations with maximum episode lengths of 120 steps. 
We  compare our proposed registration-based method, ``predictor propagation,'' and ground-truth registration obtained from the simulator, which provides an oracle upper bound on registration performance. 

\end{document}